\documentclass[conference]{IEEEtran}
\usepackage[pdftex]{graphicx}
\usepackage{multirow}
\usepackage{balance}
\usepackage[justification=centering]{caption}
\usepackage{amsmath}
\usepackage{amssymb}
\DeclareMathOperator*{\argmax}{argmax}
\DeclareMathOperator*{\argmin}{argmin}

\begin{document}
%
\title{A Procedural Texture Generation Framework Based on Semantic Descriptions}





%
\author{\IEEEauthorblockN{Junyu Dong$^\dag$,
Lina Wang$^\dag$,
Jun Liu$^\ast$,
Xin Sun$^\dag$}
\IEEEauthorblockA{$\dag$ Ocean University of China, Qingdao, China \\
$\ast$ Qingdao Agricultural University, Qingdao, China\\
dongjunyu@ouc.edu.cn, wanglina@stu.ouc.edu.cn, liujunqd@163.com, sunxin@ouc.edu.cn\\
}}


\maketitle

\begin{abstract}
Procedural textures are normally generated from mathematical models with parameters carefully selected by experienced users. However, for naive users, the intuitive way to obtain a desired texture is to provide semantic descriptions such as "regular," "lacelike," and "repetitive" and then a procedural model with proper parameters will be automatically suggested to generate the corresponding textures. By contrast, it is less practical for users to learn mathematical models and tune parameters based on multiple examinations of large numbers of generated textures. In this study, we propose a novel framework that generates procedural textures according to user-defined semantic descriptions, and we establish a mapping between procedural models and semantic texture descriptions. First, based on a vocabulary of semantic attributes collected from psychophysical experiments, a multi-label learning method is employed to annotate a large number of textures with semantic attributes to form a semantic procedural texture dataset. Then, we derive a low dimensional semantic space in which the semantic descriptions can be separated from one other. Finally, given a set of semantic descriptions, the diverse properties of the samples in the semantic space can lead the framework to find an appropriate generation model that uses appropriate parameters to produce a desired texture. The experimental results show that the proposed framework is effective and that the generated textures closely correlate with the input semantic descriptions.
\end{abstract}


%
\IEEEpeerreviewmaketitle

\section{Introduction}

Procedural textures are normally generated by computer algorithms based on mathematical models. Such models have been widely used in computer games and animations to efficiently render natural elements such as wood, marble, stone and walls. Unless a person has sufficient knowledge of procedural texture models, it is difficult to predict which model can produce what types of textures. Moreover, tuning the parameters of procedural models to produce the desired textures is a challenging task even for experienced users.

\begin{figure}[tb]
  \centering
  \includegraphics[width=8.5cm]{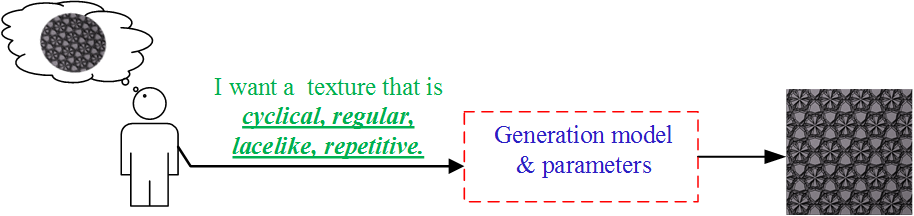}\\
  \caption{An illustrative example.}\label{fig:generation}
 \end{figure}

When users look for a procedural texture --whether for scene rendering or some other use --they will intuitively use semantic descriptions. Consider an illustrative example such as that presented in Fig.\ref{fig:generation}, one might wish to find a bark texture that is \emph{crinkled}, \emph{repetitive}, \emph{oriented} \emph{rough} and \emph{gouged}, or a granite texture that is \emph{rough}, \emph{rocky}, \emph{speckled}, \emph{dense}, \emph{regular}. Obviously, there is a large "gap" between semantic descriptions and the mathematical procedural texture models.  Therefore, the first question addressed by this study concerns how to fill this gap. The literature suggests that little work has been done concerning the retrieval of procedural textures based directly on semantic descriptions; Moreover, this is a challenging task because it involves finding both proper models and appropriate parameters.

This paper proposes a framework to bridge the gap between semantic descriptions and procedural texture models for generating desired textures. Using this framework, users can obtain a desired texture simply by providing semantic descriptions, which are used to identify a corresponding procedural model and suitable parameters for use in graphic applications. A good texture generation scheme should be able to adapt the input semantic descriptions as shown in Fig.\ref{fig:generation}. In this case, although the input description is too concise to completely express the details of the desired texture, the generated texture is well consistent with the given semantic descriptions.

To fulfill such a task, it is essential is to establish a dataset of procedural textures accompanied by their semantic attributes. Textures are particularly well suited being described in terms of attributes, as a rich lexicon has evolved for describing textures, from which appropriate terms may be readily sourced. For this purpose, we referred to the work of ~\cite{bhushan1997texture} with respect to the collection of human perceptual terms for textures, and in accordance with this work, we performed extensive psychophysical experiments to collect semantic descriptions to collect semantic descriptions for our procedural textures.
 \begin{figure}[h]
  \centering
   \begin{minipage}[b]{.9\linewidth}
      \centering
      \centerline{\includegraphics[width=5cm]{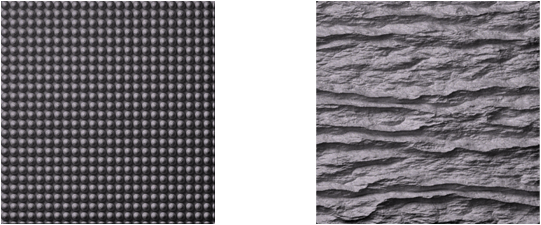}}
   \end{minipage}
  \caption{ Examples of semantic attributes that describe procedural textures. The image on the left can be described as \emph{uniform},\emph{well-ordered}, \emph{polka-dotted}, \emph{simple}, \emph{globular}, \emph{regular}, \emph{cyclical}, \emph{repetitive}; the image on the right can be described as \emph{rocky}, \emph{ridged},  \emph{crinkled},\emph{ lined},  \emph{repetitive}, \emph{rough}, \emph{irregular}.}
  \label{fig:semantic attribtutes}%
\end{figure}This process ultimately resulted in a total of 43 semantic attributes\footnote{All 43 semantic attributes are: irregular, grid, granular, complex, uniform, spiralled, marbled, mottled, fuzzy, crinkled, well-ordered, speckled, polka-dotted, repetitive, ridged, uneven, smooth, cellular, globular, porous, regular, veined, cyclical, freckled, simple, dense, stained, honeycombed, coarse, rough, gouged, rocky, woven, lined, fine, nonuniform, disordered, fibrous, random, lacelike, messy, scaly, netlike.} describing 450 textures, as exemplified in Fig.~\ref{fig:semantic attribtutes}. The core descriptions provided by these semantic attributes capture the major semantic characteristics of the samples, despite the comparatively rich visual information contained in the texture images.

\begin{figure}[h]
  \centering
   \begin{minipage}[b]{.9\linewidth}
      \centering
      \centerline{\includegraphics[width=7cm]{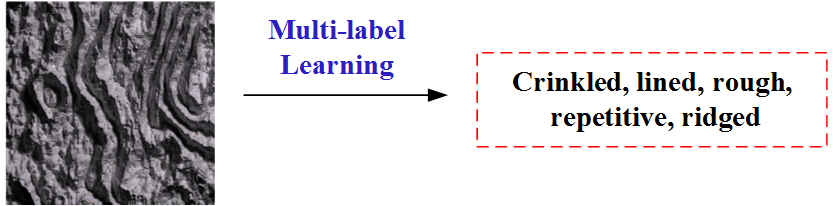}}
   \end{minipage}
  \caption{Example of the prediction of semantic attributes via multi-label learning.}
  \label{fig:example_multi}%
 \end{figure}

Before constructing the framework for generating textures from semantic descriptions, we conducted the essential and fundamental work to augment our semantic texture dataset --training a texture semantic attribute predictor. This was necessary because the number of textures with semantic descriptions collected from the psychophysical experiments was too limited to enable the construction of a robust framework. To realize this predictor, we conducted an extensive survey of research on the prediction of semantic attributes for a given image (e.g., predicting captions from natural images~\cite{Karpathy2014DeepFE}, or labeling semantic attributes for natural textures ~\cite{Cimpoi2015DeepFB}). However, because these works were unable to provide satisfactory solutions for the simultaneous annotation of a given texture ssample with multiple attributes together to achieve a full description, we selected to use a multi-label learning method, i.e., a technique that can simultaneously predict more than one label for a sample. An illustrative example of the results of multi-label learning is shown in Fig.~\ref{fig:example_multi}. Using this method, we built a relatively large semantic dataset containing 8,800 procedural textures with annotated semantic attributes from various generation models.

\begin{figure*}[tb]
  \centering
  \includegraphics[width=17cm]{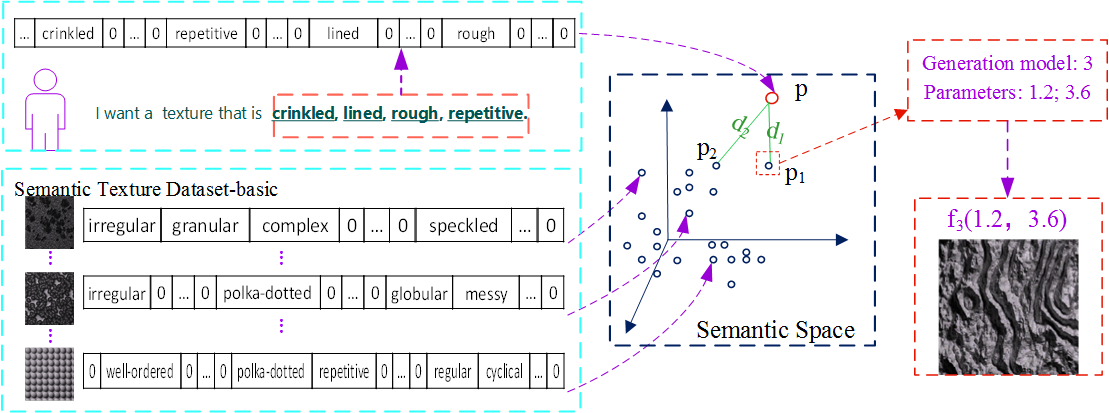}\\
  \caption{ The workflow of our texture generation process.}\label{fig:structure}
 \end{figure*}

Next, to generate textures with given semantic attributes, we constructed a semantic space based purely on the augmented semantic dataset, as Matthews et al.. did in ~\cite{Matthews2013EnrichingTA}. This was achieved by applying the manifold learning method. Note that the samples embedded into this semantic space all have an attached "tag" denoting their corresponding generation models and parameters. This is a useful addition because we can assume that the tag information is sufficient to identify the model and parameters required to generate the desired texture.
 After all this information was ready, a real desired texture's description can first be embedded into the semantic space. Then, its nearest neighbor can be found with the aid of a distance metric in this space. Finally, we employ the generation model and parameters described in the tag of the nearest neighbor to generate a new texture. This new texture image is what users really want. As shown by the example in Fig.~\ref{fig:structure}, our framework generates a procedural texture image that is simplified but coincident with the input descriptions.

The main contributions of this paper can be summarized as follows:

1. We propose a framework for generating procedural texture based on user-selected semantic descriptions.

2. We established a dataset called the basic Semantic Texture Dataset (STD-basic) for procedural texture images containing 43 semantic attributes. Compared with other datasets in state-of-the-art studies, this dataset contains more manually labeled semantic descriptions and reflects different attribute levels (i.e., the values of the attributes range from 0 to 1, meaning from "weak" to "strong")

3. To augment the semantic dataset, a multi-label method has been shown to be efficient in annotating multiple attributes for a texture image.

\section{Related Work}

Semantic information has been discussed and explored by research in image annotation and retrieval,. Some published works have investigated the correlations between visual features and semantic attributes, and proposed semantically based retrieval schemes~\cite{zhao2002bridging,Liu2007ASO,Li2007ANN}. Major studies have focused on automatic image labeling using specific visual categories or by generating natural language descriptions ~\cite{Karpathy2015DeepVA,li2010object}. Normally, such works concentrate on sentence labelling, object recognition, or region annotation. However, these studies have mainly involved natural images, not semantic attribute labelling for textures.
Because textures provide important information for describing real-world materials, texture semantic attributes can play important roles in content-based image retrieval. Certainly, there have been studies that paid attention to semantic attributes. The authors of~\cite{Datta2008ImageRI,wang2011modeling} explored high-level perceptions of textures for annotating or retrieving textile images.

When researching texture semantics, the pioneering work of Bhusan and Rao~\cite{bhushan1997texture}  identified the texture terms that humans use to describe or recognize textures and established a texture lexicon that can help people to understand the various categories of visual texture terms. Meanwhile, Cimpoi et al.. recently developed an effective texture dataset~\cite{Cimpoi2015DeepFB} that contains natural texture images labeled with 47 attributes. Unfortunately, only one key attribute was annotated by human subjects in this dataset, which means it lacks comprehensive descriptions to some degree.

By contrast, in this study, we collected all the semantic attributes for the procedural textures from psychophysical experiments as described in~\cite{bhushan1997texture}. Then, we annotated the semantic attributes for new textures as described in~\cite{geng2014multilabel}, offering a relatively more powerful approach to enrich the descriptions of procedural textures.

To understand the pertinent semantic attributes, we adopted a semantic space strategy in which the semantic space is constructed with axes corresponding to several attributes. This approach can serve as a standard representation for texture semantic descriptions. Semantic space was first proposed as a way to quantitatively describe textures in ~\cite{wang2011modeling}. Later, Matthews and Nixon ~\cite{Matthews2013EnrichingTA} also built a semantic space using hierarchical clustering to analyze textures more intuitively. In~\cite{liu2015visual}, Liu et al. discussed perceptual space and reached the conclusion that it would be more consistent with human perception if a space were built to represent the perceptual descriptions. Inspired by this work, our study aims at developing such a semantic space to represent a texture semantic descriptions and ultimately improves the similarity measure in terms of human semantic description.

Furthermore, some efforts have been devoted to bridge the gap between computational texture features and the corresponding texture semantic descriptions. These studies make use of ideas to represent texture images with high-level features. Typical ones among these computational texture features are Gabor wavelets ~\cite{Manjunath1996TextureFF,zhang1998comparison} and Local Binary Patterns (LBP)~\cite{ojala2002multiresolution}. These features have been successfully applied to texture classification and retrieval. In addition to those handcrafted features, various deep learning models, particularly deep convolutional neural networks (CNNs), have been developed for image classification and object recognition ~\cite{Krizhevsky2012ImageNetCW,zeiler2014visualizing,he2015spatial,girshick2014rich,dixit2015scene} to investigate representations with better performance. The work in~\cite{Zhao2015DeepSR}, which utilized the CNN model, explored a semantic ranking for multi-label image retrieval, but did not apply that concept to textures.

This study also compares the Gabor wavelet features and CNN features in predicting the semantic attributes for procedural textures. Furthermore, the relationship between the attributes of procedural textures and those in the semantic space are found based on a manifold learning method. In this way, automatic semantic annotation and texture generation based on semantic description are implemented for procedural models.

\section{Dataset Construction}
The first step towards our goal is to build a semantic procedural texture dataset. One convincing approach is to conduct psychophysical experiments to collect real human descriptions for these procedural textures, as in the work of~\cite{bhushan1997texture} and ~\cite{Cimpoi_2014_CVPR}. There were 450 procedural textures with a size of 512*512 in our dataset. Now we briefly review the process of acquiring semantic descriptions for these textures.
\subsection{Data Collection Based on Psychophysical Experiments}

We carried out our psychophysical experiments using the following methodology. A total of 20 people were involved in this experiment. In the initial step, the 450 procedural texture images were printed and spread out on a desktop for each of the 20 participants, making it possible for the participant to quickly scan all these images. Then each participant was asked to sort all the textures into several groups. During this step, there were no constraints concerning the number of groups; however, the participants were told that the groups should comply with the rules that textures assigned to any one group should reflect the same semantic attribute, and each group should be as detailed as possible. Subsequently, the observers wrote down the semantic attributes that the participants kept in mind during the grouping process. This semantic word selection process was not subject to any constraint for the participants; word selection was entirely based on their own personal understanding of the textures. Note that this approach may result in situations in which different semantic words from different participants with the same meaning describe the same texture. In such cases, we later selected one such word as the ultimate attribute for data processing purposes. At the end of this process, we had recorded the semantic attributes and their corresponding textures.

After the first grouping step, each texture had been assigned one core semantic attribute by each participant. To enrich every texture with additional semantic attributes, the participants were asked to continue the experiment by merging some of the groups they had created in the first step. This requires the participant to select groups that can reflect a more general semantic attribute and merge them. For example, suppose one of group has the attribute "lined," while another has the attribute "globular." A participant might merge these two groups into one cluster because they both can be subsumed under the attribute "repetitive". After this operation, another more general semantic attribute can be added to each texture. To take uncertainty into account, we asked the participants to mark down their confidence level (as a value between 0 and 1). Later, this value would be considered in data processing step during every merging operation. To apply even more general attributes describing the textures, the participants were requested to repeatedly merge the clusters from the last merging process until there only two clusters remained. All these general attributes were recorded during the merging process.

When one participant completed the task early while other people were still occupied, they were asked to repeat the aforementioned steps.

\subsubsection{Data processing} 

The next step was to process the data obtained from the experiments described above. For the attributes, two operations were conducted. First, we selected as a standard one of the words with the same meaning as the ultimate attribute as in~\cite{bhushan1997texture}. Second, from all the descriptive attributes we selected those words for our semantic descriptions that were most meaningful and those in which the participants had the highest confidence based on the 98 words proposed in~\cite{bhushan1997texture}. We discarded terms that did not make sense or that had low confidence values. Finally, we ended up with 43 semantic attributes describing the 450 textures in the dataset.

For each of the 43 semantic attributes of each texture, we counted how many of the 20 people assigned it to each attribute. This number represented the intensity of the relationship between that texture to the attribute. For example, if 5 out of the 20 participants assigned texture \emph{x} to the attribute "regular", then texture \emph{x} would be labeled with a value of 5 for the "regular" attribute. The number 5 then is divided by 20, representing the value of "regular" for texture \emph{x}. By following this process, we finished building the semantic texture dataset.

To further expand the dataset with real attributes, we divided each 512*512 texture into 4 samples, finally obtaining 1,800 textures with 43 semantic attributes as the final basic Semantic Texture Dataset (STD-basic).

\subsection{Prediction of Texture Semantic Attributes}

As discussed above, it is necessary to augment our STD-basic dataset because it contains only 1,800 texture images with semantic attributes, which is somewhat deficient for constructing a framework to generate textures from human descriptions. Hence, we selected an additional 8,800 procedural textures to augment our semantic texture dataset. These were produced by dividing 2,200 textures with a size of 512*512 into 4 parts. We refer to this second dataset as the additional Semantic Texture Dataset (STD-add). The textures in STD-add were generated using the same generation models as the textures in STD-basic, but were rendered under different illumination condition.

To augment the new dataset with the semantic attributes, we investigated multi-label learning, a technique that allows samples to have more than one label simultaneously. We briefly introduce the fundamentals of multi-label learning techniques below.

From the perspective of data analysis, it is reasonable to assume that each human subject labels textures with semantic attributes according to a latent distribution with respect to specific textures, whether consciously or unconsciously. Therefore, we can assume that the semantic values of these texture images also follow a latent semantic distribution overall. In contrast, given a texture instance \emph{x}, the goal of multi-label learning method is to predict several labels for \emph{x} that can satisfy the latent distribution hidden in the semantic attributes as much as possible~\cite{geng2014multilabel}. This problem can be solved via two steps. The first step is to generate a common distribution for each instance that is most compatible with the latent distribution. The second step is to learn a mapping from the instance space $\chi$ = R$^{q}$  to the latent distribution space. Then, we could implement our semantic attribute predictor after training the multi-label learning model.
\subsubsection{Modeling the Semantic Distribution}

Motivated by the multi-label learning method described above, we explored the distribution of semantic attributes in the STD-basic dataset. First, we generated a semantic distribution \emph{P}  that is most compatible with the average values of the semantic attributes assigned to the textures. Then we learned this distribution to estimate the values of semantic attributes for additional texture images. Here, the learned distribution \emph{P} works similarly to an attribute predictor. This distribution P from STD-basic should satisfy the semantic values$\emph{P}_{i}$ of the\emph{i}-th texture image $\emph{x}_{i}$ as much as possible:
 \begin{equation}\label{eq:1}
   P = argmin\sum_{i=1}^{n}D(P,P_{i})
    \end{equation}
where D is a function measuring the distance/similarity between the two distributions. In addition, both \emph{P} and $\emph{P}_{i}$ should be constrained to obey $\emph{P}_{i}^{\emph{j}}\geq0$, $\sum_{\emph{j}}\emph{P}_{\emph{i}}^{\emph{j}} = 1$, where $\emph{P}_{i}^{\emph{j}}$ is the value of \emph{j}-th semantic attribute for \emph{i}-th sample, and \emph{j} = 1,\dots,\emph{c}, \emph{i} = 1,\dots,\emph{n}, \emph{c} is the number of semantic attributes, and \emph{n} is the number of texture samples.

Many choices for defining D in Eq.(\ref{eq:1}) have already been studied~\cite{cha2007comprehensive}, such as the commonly used Euclidean distance, Kullback-Leibler(KL), Squared $\chi^{2}$, or the Intersection and Fidelity similarity measures. In this study, we applied the KL divergence and formulated the learning problem as the following nonlinear optimization process:
\begin{eqnarray}\label{eq:1-1}
 \min &~&~\sum_{i=1}^{n}\sum_{j=1}^{c}\emph{P}^{j}\log\frac{\emph{P}^{j}}{\emph{P}_{\emph{i}}^{\emph{j}}}  \\ \nonumber
   \emph{w.r.t.}&~&~\emph{P}^{\emph{j}},~\emph{P}_{\emph{i}}^{\emph{j}},~\emph{j} = 0,\dots,c, \emph{i} = 1,\dots,n  \\ \nonumber
     \emph{s.t.}&~&~\emph{P}^{j}\geq0, \emph{P}_{i}^{\emph{j}}\geq0  \sum _{\emph{j}}\emph{P}^{\emph{j}} = 1, \sum_{\emph{j}}\emph{P}_{\emph{i}}^{\emph{j}} = 1.
       \end{eqnarray}
This progression is applied to each training sample $\emph{x}_{i}$, resulting in the semantic distribution \emph{P} that incorporates the probability distributions of all the instances.

\subsubsection{Learn the Semantic Distribution}

After establishing the distribution \emph{P}, we need to learn this distribution to predict our semantic attributes. Generally speaking, we have transformed our training set into this form: $\emph{X}={(\emph{x}_{1}, \emph{P}(\emph{x}_{1})), \dots, (\emph{x}_{n},\emph{P}(\emph{x}_{n}))}$, where \emph{P}(\emph{x}$_{i}$) is the semantic distribution associated with the \emph{i}-th sample \emph{x}$_{i}$. The main objective is to learn a conditional probability parametric model \emph{p}(\emph{y}$|$\emph{x};$\theta$) from the training set, where $\theta$ is the parameter vector. Therefore, the problem to learn this distribution becomes one of learning the parameter $\theta$  that causes the distribution \emph{P}(\emph{x}$_{i}$) to be most similar to the semantic distribution on sample \emph{x}$_{i}$. The best compatible parameter vector, $\theta^{*}$, is defined as follows:

\begin{align}
 \label{eq:2}  \theta^{*} &= \ \  {\argmin}\sum_{i}\sum_{j}(v_{x_{i}}^{y_{j}}\ln\frac{v_{x_{i}}^{y_{j}}}{p(\emph{y}_{j}|\emph{x}_{i};\theta)}) \\
   \label{eq:23}             &= \ \  {\argmax}\sum_{i}\sum_{j}v_{x_{i}}^{y_{j}}\ln p(\emph{y}_{j}|\emph{x}_{i};\theta).
     \end{align}

Similar to ~\cite{geng2013facial}, we assume that the conditional probability fits the maximum entropy model~\cite{berger1996maximum}:

\begin{equation}\label{eq:3}
   p(\emph{y}_{j}|\emph{x}_{i};\theta) = \frac{1}{Z}\exp(\sum_{k}\theta_{y,k}\emph{x}^{k}),
     \end{equation}

where $Z = \sum_{y}\exp(\sum_{k}\theta_{y,k}\emph{x}^{k})$ is the normalization factor, $\theta_{y}$ is an element in $\theta$, and $\emph{x}^{k}$ is the \emph{k}-th attribute of instance \emph{x}. Substituting Eq. (\ref{eq:3}) into Eq. (\ref{eq:23}) yields the target function

\begin{align}
   \emph{T}(\theta) &= \sum_{\emph{i,j}}v_{x_{i}}^{y_{j}}\sum_{k}\theta_{y_{j},k}\emph{x}_{i}^{k} \\
    \label{eq:4}    &- \ \  \sum_{i}\ln\sum_{j}\exp(\sum_{k}\theta_{y_{j},k}\emph{x}_{i}^{k}).
      \end{align}

The minimization of $\emph{T}'(\theta) = -\emph{T}(\theta)$ can be effectively solved resorting to the quasi-Newton method BFGS~\cite{nocedal2006numerical}. Once the distribution model \emph{p}(\emph{y}$|$\emph{x};$\theta$) has been learned, the semantic distribution of an given sample \emph{x}$_{i}$ can be predicted by it, then more and more texture instances can be enriched with semantic attributes.

\section{Texture Generation Based on Given Semantic Descriptions}

The next step is to discuss the generation process based on semantic descriptions. Intuitively, as shown in Fig ~\ref{fig:generation}, the concept is to call the mathematical models to produce a texture in accordance with given semantic descriptions. However, the critical problem is that procedural texture generation not only requires a model but also requires corresponding parameters. For naive users, the types of textures that any specific set of parameters might produce is unpredictable. Hence we explored a novel idea: by assuming that we can find the most similar texture to the given semantic descriptions in the STD-add dataset, we can use the parameters of this found texture and use those to call the generation models to generate a new texture.

In this stage, the most difficult problem is to find the most similar texture to the given semantic description. Another problem is how to use the texture parameters in STD-add. For the first problem, if we measure the similarity in the original space between the semantic descriptions directly, we will obtain a similarity matrix as in Fig.~\ref{fig:cor_pdist}, however, that is too sparse to achieve good results. Hence, motivated by previous work in~\cite{bhushan1997texture}, which mapped the original perceptive descriptions into a perception space to analyze the relationships between these descriptions, we also derived a space, one based on STD-add, in which we can separate the semantic descriptions from each other. We call this space the Semantic Space because it was derived from the 43-dimensional semantic descriptions. In this semantic space, we can subsequently conduct a nearest neighbor retrieval to find the example most similar to the given semantic description. Eq.(\ref{eq:10}) provides a generalized way to perform the nearest neighbor retrieval. In this approach \emph{x} is a projection point in the semantic space of the tested semantic vector that describes a desired texture, \emph{X} stands for all the samples in our semantic space, \emph{D} is a general distance metric method to find the most similar semantic description (for example, it could be the Euclidean distance, Manhattan distance, or the Chebyshev distance―here, we adopted the Euclidean distance).

\begin{equation}\label{eq:10}
  L(x) =  \min D(x,X)
   \end{equation}

For the second problem, the textures in STD-add all have an annotating tag indicating their own parameters, including which model they were generated from and what parameters were used. Based on this tag, after finding the most similar semantic description, we simply take both the generation model and the corresponding arguments from the attached tag to generate a new texture.

More specifically, taking Fig.~\ref{fig:structure} as an example, $\emph{p}_{1}$ is a semantic description's projection in semantic space that is tagged with its own generation parameters, indicating that it was generated by the third generation model using the parameters 1.2 and 3.6, as shown in the upper right of the image. Now, when a user provides semantic descriptions of a desired texture, we can project the description vector into a point \emph{p} in the semantic space, as shown in Fig.~\ref{fig:structure}. Then we can find its most similar point $\emph{p}_{1}$ and use the tagged parameters to generate a new texture. Fig.~\ref{fig:structure} exemplified our entire procedure for generating new textures.

As introduced in the procedure above, the semantic space is vitally important in generating new textures. Next we describe how this space was built in more detail.

\subsection{The Semantic Space}

To create the semantic space, we resort to the isometric feature mapping algorithm (Isomap). Theoretically, the goal of Isomap is to find a low dimensional embedding that can maintain the neighborhood structure between data points in the high dimensional manifold, and it is advantageous because the natural dimension \emph{d} of the low dimensional embedding can be determined by the residual error ~\cite{tenenbaum2000global}.

\begin{figure}[tb]
      \centering
       \includegraphics[width=9cm]{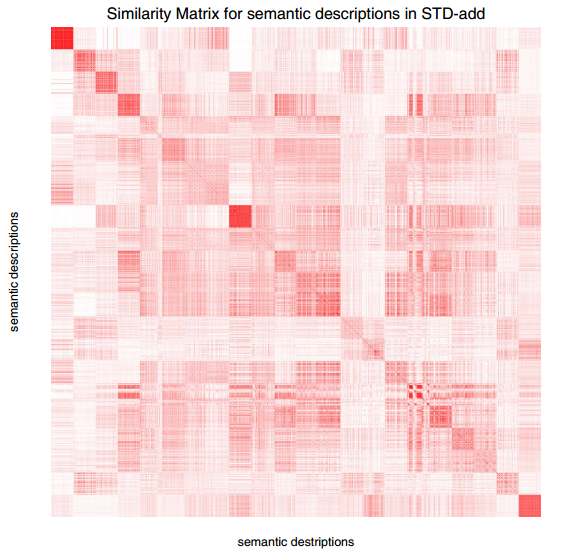}\\
        \caption{The similarity matrix of 8,800 textures. The color variations from white to red represent the degree of similarity between pairs of samples in terms of semantics (i.e., from the least similar to the most similar, respectively).}\label{fig:cor_pdist}
     \end{figure}

First, we computed the similarity matrix from the semantic descriptions in STD-add (which reasonably was the pairwise correlation coefficients of all semantic descriptions in STD-add) to act as the input distance matrix for the Isomap algorithm. For clarity, this concept is illustrated in Fig.~\ref{fig:cor_pdist}; the color variations from white to red represent the degree of similarity between pairs of samples in terms of semantics (i.e., from the least similar to the most similar). Because all the textures in STD-add were generated from 22 generation models, there are, generally, 22 red areas that indicate more similar pairs shown on the diagonal. Fig.~\ref{fig:cor_pdist} also expresses the information that, on the whole, textures generated from the same generation model are more similar to each other than they are to those from different generation models.

\begin{figure}[tb]
      \centering
       \includegraphics[width=9cm]{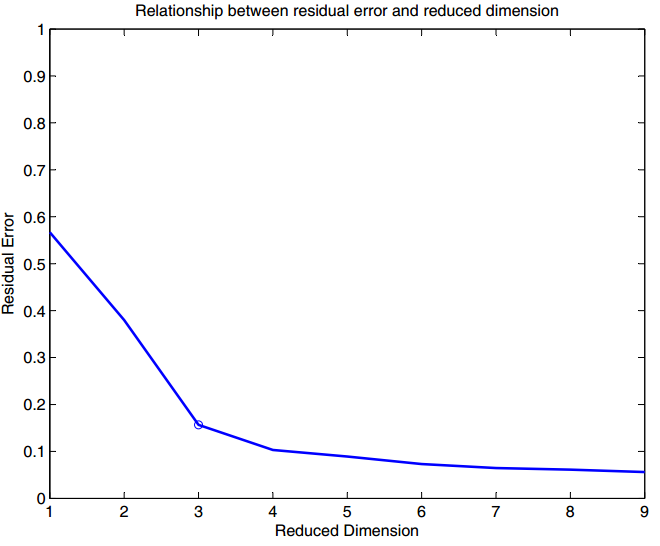}\\
        \caption{Relationship between the reduced dimensions and the residual error.}\label{dimension_residural error}
     \end{figure}

\begin{figure}[tb]
    \centering
     \includegraphics[width=9cm]{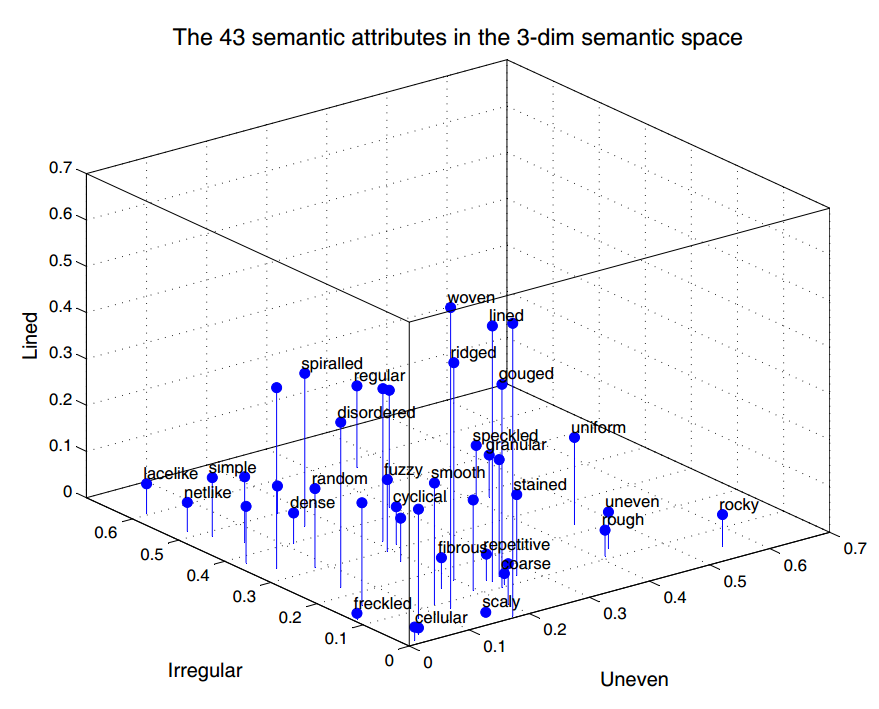}\\
       \caption{The correlations between the 43 semantic attributes and the three coordinates of the semantic space.All 43 semantic attributes are annotated with their approximate locations.}\label{fig:XYZ}
        \end{figure}

As mentioned earlier, Isomap's main advantage is that the natural dimension \emph{d} of the low dimensional embedding can be determined by the residual error. Thus, we investigated the residual errors under different dimensions. Fig.~\ref{dimension_residural error}, which shows how the curve of residual error changes with different dimensions, shows that there is an inflection point at \emph{d}=3, where the curve changes significantly. Thus \emph{d}=3 can certainly be considered as the intrinsic dimension of our semantic space. Furthermore, we explored the relationship by computing the correlation coefficients between 43 semantic attributes in 3-dimensional semantic space and showed them in Fig.~\ref{fig:XYZ}.

Fig.~\ref{fig:XYZ} clearly plotted the correlations semantic attributes and each of the coordinates in the semantic space. We selected those attributes whose correlation coefficient was greater than 0.55 on coordinates X, Y, Z and found that \emph{’irregular, complex, spiraled, fuzzy, well-ordered, porous, regular, veined, cyclical, simple, dense, honeycombed, nonuniform, random, lacelike, netlike’} are more related to X coordinate; \emph{’granular, uniform, mottled, speckled, repetitive, uneven, cellular, stained, coarse, rough, rocky, fibrous, scaly’} are more related to Y coordinate; and \emph{’grid, marbled, crinkled, polka-dotted, ridged, smooth, globular, gouged, woven, lined, fine, disordered, messy’} are more related to Z coordinate. It can be seen that attributes related to the X coordinate have the common attribute "irregular", while those related to Y have the common attribute "uneven", and those to Z have the common attribute "lined".

The choice to build this semantic space was the key to generate textures from semantic descriptions. Hypothetically, if we had not built such a space for the 43-dimensional semantic descriptions, it would have been far more complicated to find the generation parameters to generate a desired texture. However, by building this semantic space, where samples are adequately separated from each other, we can build a bridge to fill the gap described in the Introduction.

\section{Experimental Results}

We can now turn to a practical evaluation of generating textures in accordance with given semantic descriptions. Considering the dataset augmentation procedure, we begin with an experiment to evaluate semantic attributes prediction and then introduce the results of generating textures.

\subsection{Data Augment}
In Section 2.2 we discussed the overall procedure to predict semantic attributes for textures. Here, we will provide the details of how we can predict semantic attributes for textures.
\subsubsection{Texture Representation}

To support the theoretical statements in Section 2, we provide measurements and comparisons by selecting different representations for texture images. The selected pertinent features are as follows:
  \begin{itemize}
   \item The Gabor wavelet responses of 24 orientation and scale combinations given in~\cite{Manjunath1996TextureFF}, yielding a feature vector of 48 elements.
   \item The 4096-dimensional CNN features directly extracted from the penultimate fully-connected layer of the CNN model AlexNet~\cite{Krizhevsky2012ImageNetCW}, which contains 8 layers: 5 convolutional layers and 3 fully connected layers.
  \end{itemize}

Because the Gabor features have been demonstrated to be reasonably efficient at representing textures, and the CNN features have recently achieved great success in many fields, we chose these two feature sets to predict semantic attributes.
\begin{figure}[tb]
  \centering
   \includegraphics[width=8cm]{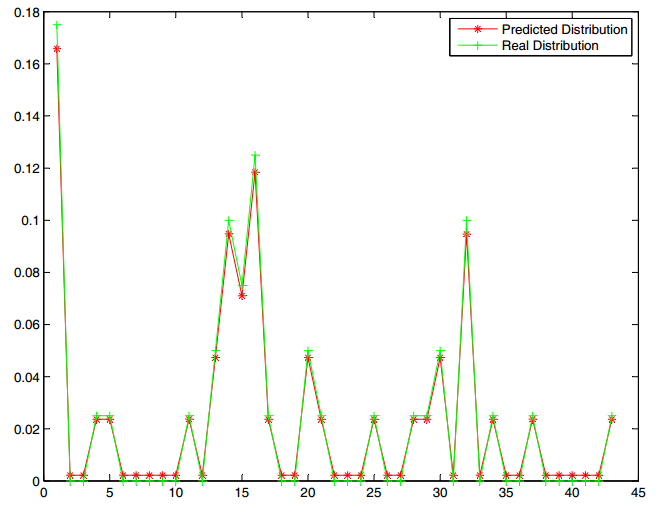}\\
    \caption{Comparison of the 43 semantic attributes predicted using Gabor features with the real semantic attributes obtained from the psychophysical experiments. The red line represents the predicted distribution for the test textures, and the green line represents the real distribution of the semantic attributes.}\label{fig:se_pre_gabor}
       \end{figure}

  \begin{figure}[tb]
    \centering
      \includegraphics[width=8cm]{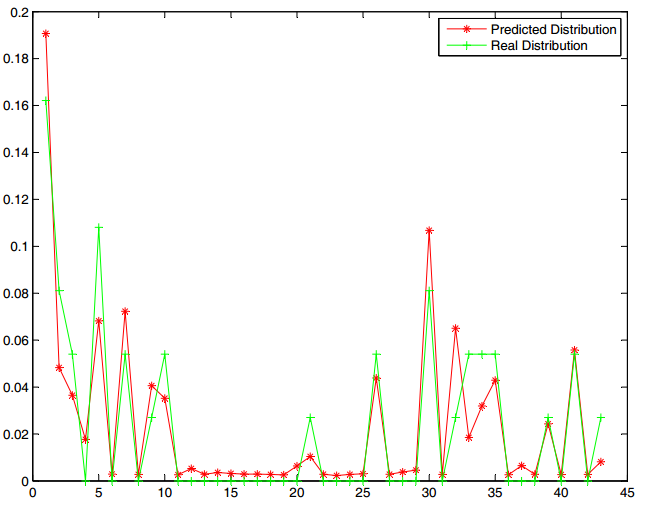}\\
        \caption{Comparison of the 43 semantic attributes predicted using CNN features with the real semantic attributes obtained from the psychophysical experiments. The red line represents the predicted distribution for the test textures, and the green line represents the real distribution of the semantic attributes.}
           \label{fig:se_pre_CNN}
           \end{figure}

\subsubsection{Semantic Attributes Prediction}

With the two kinds of descriptors representing textures, the multilabel learning method discussed in Section 2 was applied to predict semantic attributes. The parameters of the algorithm were set the same as described in~\cite{geng2014multilabel}, except for the penalty coefficient, \emph{C}.

The performance of the multi-label learning method was evaluated on the STD-basic dataset, from which 1,700 samples were used to train the prediction model and the rest of the samples were used for testing. The prediction results derived from the models trained on the Gabor features and the CNN features are shown in Fig.~\ref{fig:se_pre_gabor} and ~\ref{fig:se_pre_CNN}, respectively.

Interestingly, Fig.~\ref{fig:se_pre_gabor} and ~\ref{fig:se_pre_CNN}, seem to show that the prediction using the Gabor features achieves better results than the prediction using the CNN features. The multi-label learning method can automatically fit the distributions of semantic attributes under the Gabor features while it appear relatively large fluctuations when using the model trained on the CNN features. Furthermore, Table.~\ref{table:seGabor} lists the distance between the predicted distributions and the original distributions of the semantic attributes under both the Gabor and CNN features. These results suggest that the predicted distributions by Gabor features are closer to the original distributions than those predicted by the CNN features. Intuitively, one might expect better results from the CNN features than the Gabor features as the CNN features have shown powerful representation abilities for classification purposes and in many other areas~\cite{zeiler2014visualizing,he2015spatial,girshick2014rich,dixit2015scene}. The inferior performance of CNN features in this study may have occurred because the AlexNet model was trained on the ImageNet dataset, whose samples vary from animals to planes, while the textures in the STD-basic dataset are quite different from those(as Fig.~\ref{fig:semantic attribtutes} shows). Meanwhile, the Gabor features have been shown to be one of the most suitable feature sets for representing textures; therefore it is not that surprising the Gabor features achieved a better result under the study conditions.

    \begin{table}
        \begin{center}
        \begin{tabular}{|l|c|c|c|c|c}
        \hline
        \  \   &kldist & euclideandist & sorensendist &$\chi^{2}$ dist \\ \hline 
        \  \  Gabor & 0.0564 & 0.0195 & 0.0557 &0.0564\\ \hline 
        \  \ CNN & 0.1114 & 0.0671 & 0.1380 &0.1095\\ \hline 
        \end{tabular}
        \end{center}
        \caption{Average distances between the real semantic attributes obtained from the psychophysical experiments and the semantic attributes predicted using Gabor features and CNN features. Four distance measures were used: the KLD distance, the Euclidean distance, the Sorensen distance and the $\chi^{2}$ distance.}\label{table:seGabor}
        \end{table}

Consequently, in this study, the Gabor features are more suitable than the CNN features for predicting semantic attributes using the multi-label learning method. Therefore, with the Gabor features representing textures in STD-add, the semantic attributes for STD-add can be predicted by the model trained on the Gabor features.

\subsection{Procedural Texture Generation Based on Semantic Descriptions}

\begin{figure*}
\begin{center}
 \centering
  \includegraphics[width=16cm]{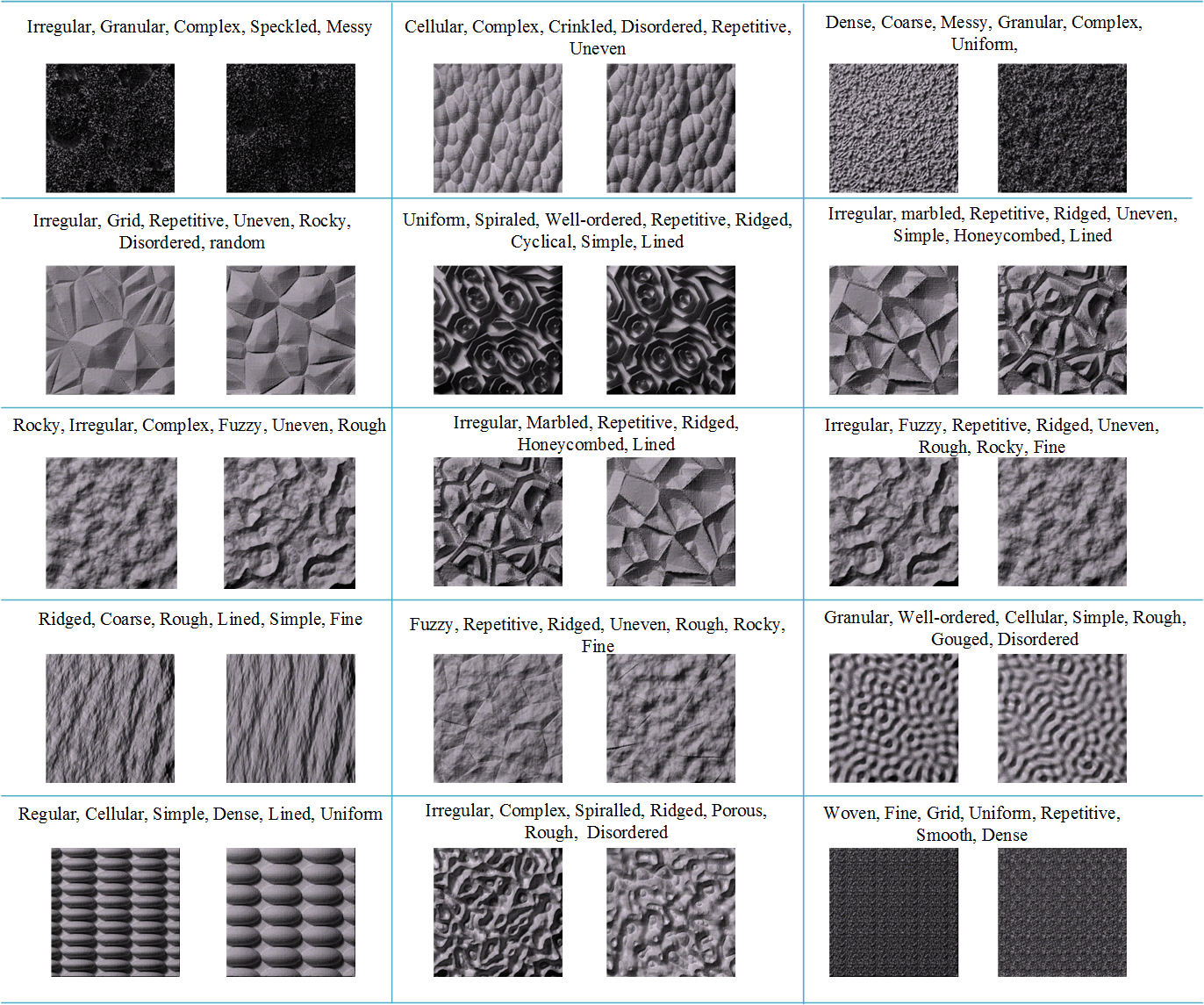}\\
   \end{center}
    \caption{ Examples of textures generated based on specified semantic attributes. The terms presented above each pair of textures constitute the input semantic vector. The texture on the left is the new texture generated according to the input semantic vector, and the texture on the right is the nearest-neighbor sample found in the semantic space.}\label{fig:generation1}
    \end{figure*}
We follow the procedure described in Section 3 to find the appropriate generation model and its parameters in order to generate a new texture in accordance with a given semantic description. Fig.~\ref{fig:generation1} presents some of the generation results for different semantic descriptions. Heterogeneous descriptions were tested during the generation process, and the generated textures did conform to the input semantic attributes. The results in Fig.~\ref{fig:generation1} verified the effectiveness of the proposed method to generate textures in accordance with semantic descriptions.

Meanwhile, to measure the generation performance, we tested some of the semantic descriptions from STD-basic because that dataset was comprehensively labeled with ground-truth data. The experimental results show that the mean square error (MSE) between the input semantic description vector and the predicted semantic vector of newly generated textures is 0.0246. This small number offers strong evidence for the validity and reliability of this texture generation method.

\section*{Conclusion}
In this paper, we proposed a novel framework to generate procedural textures matching a given semantic description. We established a dataset named STD-basic in which the contained textures were labeled with 43 semantic attributes based on psychophysical experiments. These semantic attributes were then used to label additional textures by using a multi-label learning method, resulting in a second dataset named STD-add dataset. Then by building a semantic space, we were able to bridge the gap between the semantic descriptions and texture generation models. In practice, our framework can automatically generate a procedural texture based on a set of given semantic attributes. In future work, we plan to focus on training generative models to generate textures freely, without having to search for parameters.

%

\balance
\bibliographystyle{IEEEtranS}

\end{document}